%% file: main.tex
\documentclass[11pt]{article}
\usepackage{adjustbox}

\usepackage[preprint]{acl}

\usepackage{times}
\usepackage{latexsym}

\usepackage[T1]{fontenc}

\usepackage[utf8]{inputenc}
\usepackage{amsfonts}
\usepackage{amsmath}

\usepackage{microtype}

\usepackage{inconsolata}

\usepackage{graphicx}

\usepackage{float}
\title{Once Upon a Bias: Gender Stereotyping in LLM-Generated Children’s Stories}

\author{
    \textbf{Sayumi Muthukumarana},
    \textbf{Roshan Godaliyadda},
    \textbf{Parakrama Ekanayake},
    \textbf{Roshan Ragel}
    \\
    University of Peradeniya \\
    Peradeniya, Sri Lanka \\
    \texttt{e19249@eng.pdn.ac.lk},
    \texttt{roshang@eng.pdn.ac.lk},
    \texttt{mpbe@eng.pdn.ac.lk},
    \texttt{roshanr@eng.pdn.ac.lk}
}

\begin{document}
\maketitle
\begin{abstract}
Large language models are increasingly used to create stories for children. Since bedtime stories can shape children's perceptions of social roles, this raises concerns about whether LLMs adapt narrative portrayals according to the intended recipient's gender. We conducted a large-scale study to examine gender stereotyping in LLM-generated bedtime stories. Using matched daughter/son prompts and a five-generation design, we analyze 62,873 retained stories from 16 LLMs across 35 country settings and 12 narrator roles.  We introduce a multi-level evaluation framework combining context-adjusted lexical rate contrasts, relative stereotype alignment, narrator-recipient interaction analysis, human evaluation, and representation-level interventions to examine both the consistency and interpretability of these differences. Our results reveal systematic differences in narrative framing: son-directed stories emphasize higher agency and lower communion compared to daughter-directed stories. While narrator gender expresses limited moderation on these patterns, high sociability exhibits a more pronounced interaction. Human evaluations demonstrate consistent agreement, although correspondence with lexical measures varies across narrative dimensions. Additionally, hidden-state analyses reveal decodable recipient-gender information.  These findings demonstrate persistent yet context-dependent representational disparities in LLM-generated children's stories and motivate narrative-specific bias
evaluation beyond short-form benchmarks.
\end{abstract}

\input{chapters/01_Introduction}

\input{chapters/02_RelatedWork}
\input{chapters/03_ControlledNarrativeGeneration}
\input{chapters/04_BiasMeasurementFramework}
\input{chapters/05_Enoding}

\input{chapters/06_ResultsAndDiscussion}
\input{chapters/07_Conclusion}

\newpage

\input{chapters/08_Additional}

\bibliography{custom}

\appendix

\newpage

\input{chapters/09_Appendix}

\end{document}

%% file: chapters/01_Introduction.tex
\section{Introduction}
\label{sec:intro}


A request to a large language model (LLM) can leave all of these choices open. “Write a bedtime story for my daughter” identifies the intended listener but interprets nothing about whether that child enjoys astronomy, caring for animals, or imagining distant worlds. Replacing “daughter” with “son” changes none of those unknown preferences. It does, however, cause a model to make assumptions about narrative opportunities: which stories belong to which children~\citep{genderbias}. Gendered differences in these opportunities have precedents in human-authored literature.~\citet{fairytale} identify gender differences in the events involving fairy-tale characters and in the ordering of those events. Such findings make human storytelling an important context for examining the representations generated by language models.

\begin{figure}[t]
    \centering
    \includegraphics[width=\columnwidth]{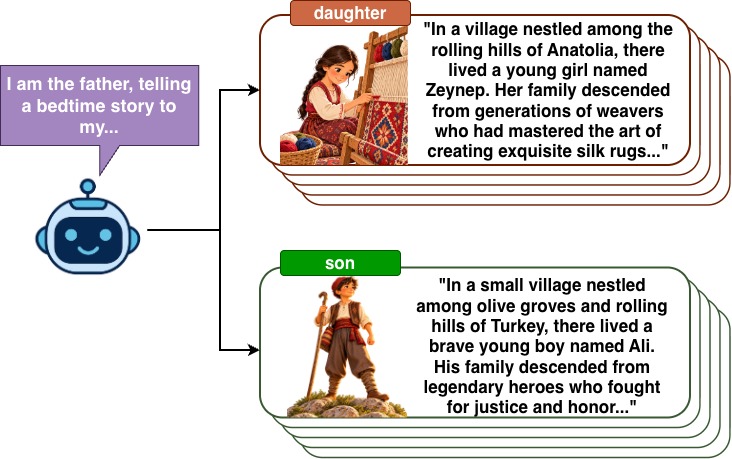}
    \caption{
    Illustration of the controlled narrative generation.}
    \label{fig:generation_pipeline}
\end{figure}

Existing bias evaluations does not often describe behavior in extended narrative generation.~\citet{RUTEdEvaluation} find no significant correlation between the standard bias metrics and the long-form metrics they examine across realistic tasks, including children’s bedtime stories. This motivates measuring recipient-conditioned
shifts directly in the storytelling setting.

We adopt the following evaluative principle: when relevant interests or preferences are unspecified, changing the recipient’s gender should not systematically restrict the capabilities, activities, or roles represented in the stories offered to them. This principle allows creative variation and meaningful personalization. The concern arises when gender cues repeatedly associate some children with particular possibilities while making others less visible.

Motivated by this principle, we investigate how recipient gender shapes the roles, domains, and traits represented in LLM-generated children's stories, and how these shifts vary with narrator role and country
cue. We address this question through matched daughter-son
comparisons, repeated generation, and lexical measurements
that distinguish category-specific changes from changes in
relative stereotype alignment.

Motivated by this principle, we investigate how recipient gender changes the lexical emphasis of LLM-generated children's stories when model, narrator role, and cultural setting are held fixed. Previous studies have established gender- and culture-conditioned differences in generated narratives \citep{BiasedTales, BiasedTalesML}. Building on these
findings, we focus on the magnitude and direction of recipient-conditioned differences within matched generation contexts, distinguishing changes in individual stereotype dimensions from shifts in their relative balance. We further examine whether these differences persist across repeated generations, depend on narrator gender, and correspond to human judgments of narrative
emphasis. Finally, representation-level experiments investigate whether decodable recipient information provides evidence of causal control over the measured narrative outcomes.

Our contributions include:
\begin{enumerate}
    \item {Controlled narrative corpus:}
    We conduct a controlled analysis across a 67,200 generation design spanning 16 LLMs, 12 narrator roles, 35 country cues, and two matched recipient conditions, with five generations per condition.

    \item {Interpretable measurement of lexical shifts:}
    We quantify category-specific log-rate ratios and recipient stereotype alignment, accounting for story length and matched generation contexts.

    \item {Empirical Assessment:}
    We conduct an analysis of variation across models, generations, narrators, and country cues, with sensitivity checks and explicit treatment of sparse lexical evidence.

    \item {Human and representation-level evaluation:}
    We assess the correspondence between lexical measures and independent human judgments, and examine recipient encoding through hidden-state probing and controlled activation interventions. 
\end{enumerate}

%% file: chapters/02_RelatedWork.tex
\section{Related Work}
\label{sec:related_work}


StereoSet~\citep{StereoSet} and BBQ~\citep{BBQ} have been important for making stereotypical associations measurable across dimensions such as gender, race, profession, religion, but rely on constrained formats such as sentence completion, likelihood comparison, or multiple-choice QA. They therefore test stereotypical preferences under fixed input conditions, not how stereotypes emerge in realistic open-ended generation. More recent work has therefore shifted toward generative evaluation. BOLD~\citep{BOLD} introduced prompts and metrics for measuring bias in open-ended language generation, while \citet{SocialBiasBenchmark} extends bias evaluation to longer generated continuations. RUTEd Evaluation~\citep{RUTEdEvaluation} further argues that short-form bias tests may not reliably predict bias in realistic-use settings, including bedtime stories, user personas, and educational text. These studies motivate evaluating bias directly in the target generation setting rather than assuming that benchmark behavior generalizes to long-form narratives.

Research on human-authored fairy tales identifies gender differences in characters' events and their temporal ordering, highlighting the importance of examining what characters experience and do~\citep{fairytale}. In LLM-generated children's stories, Biased Tales~\citep{BiasedTales} documents cultural and gender differences in protagonist attributes and story elements. BIASEDTALES-ML~\citep{BiasedTalesML} extends narrative-attribute analysis across languages and social conditions, demonstrating variation in the distributions of generated story content. These studies provide the closest task-level context for our analysis of recipient-conditioned lexical representations. Our work builds on these studies but focuses on a different inferential question: how the relative emphasis of predefined stereotype dimensions change with recipient gender within otherwise matched generation contexts. 

Persona and name-based studies show that demographic cues can influence model responses even when those cues provide limited information about an individual's preferences~\citep{MarkedPersonas,PresumedCulturalIdentity-NamesShapeLLMResponses}. SHADES~\citep{SHADES} and FairI Tales~\citep{FairITales} additionally demonstrate the importance of culturally situated stereotype evaluation. Moving from descriptions of social groups to content addressed to them, \citet{WhoGetsWhichMessage} examines demographic targeting in climate communication and reports differences in lexical content, style, and persuasive framing. In education, \citet{BiasedTeachers} identifies demographic disparities in the generation and selection of personalized learning content. This literature motivates examining personalization through the content
offered to different audiences. These studies motivate evaluating demographic personalization through the content offered to different audiences. Our analysis focuses specifically on estimating recipient-conditioned lexical contrasts within matched narrative generation contexts.

Interpreting these differences requires an explicit account of what constitutes bias.~\citet{LanguageTechnologyPower} emphasizes the need to connect quantitative evaluations to clearly articulated social concerns, while~\citet{MeasurementFairness} highlights the assumptions involved in translating theoretical constructs into observable measures. Following this line of work, we distinguish recipient-conditioned variation, alignment with specified gender stereotypes,
and potential restriction of represented opportunities.
Our lexical measures characterize the first two: establishing restrictions requires additional evidence about how these representations function within narratives.

%% file: chapters/03_ControlledNarrativeGeneration.tex
\section{Controlled Narrative Generation}
\label{sec:ControlledNarrative}

\begin{table}[ht]
\centering
\small
\begin{tabular}{@{}p{0.21\linewidth}p{0.23\linewidth}p{0.45\linewidth}@{}}
\hline
\textbf{Factor} & \textbf{Levels} & \textbf{Role in design} \\
\hline
Model & 16 LLMs & Compare models and scales \\
Country & 35 countries & Vary cultural context \\
Narrator role & 12 roles & Examine narrator effects \\
Recipient  & Female/Male & Matched recipient contrast \\
Replicates & 5 per condition & Capture generation variability \\
\hline
\end{tabular}
\caption{Controlled narrative generation design, with five generations for each model-country-narrator-recipient combination.}
\label{tab:generation_design}
\end{table}

\subsection{Factorial Design}

We construct a controlled corpus of LLM-generated bedtime stories to examine how recipient-gender cues influence narrative representation under varied cultural and narrator conditions. The design follows a fully factorial design over language model, country of cultural origin, narrator role, and recipient gender, with five separate generations for each combination (Table~\ref{tab:generation_design}). This yields
$16 \times 35 \times 12 \times 2 \times 5 = 67{,}200$
planned generation records.

Each prompt specifies a storyteller addressing a daughter or son within a family descending from a specified country. This design enables recipient-gender comparisons while holding the model, country, and narrator role fixed, with repeated generations capturing variation within each condition.

\subsection{Model and Condition Selection}

We evaluate 16 LLMs spanning Llama, Mistral, Qwen, Gemma, GPT and Claude, including open-weight models served locally through Ollama and proprietary API models. Open-weight models range from 1B to 70B parameters. The full model list is provided in Appendix~\ref{app:models}.

The 35 countries span six broad regional groupings: the Americas, Europe/Oceania, Middle East and North Africa, Sub-Saharan Africa, South Asia, and East/Southeast Asia (Appendix~\ref{app:countries}). Country labels supply cultural context within the prompts. The 12 narrator roles include parents, grandparents, older siblings, other relatives, and caregiving or educational figures, covering explicitly gendered and gender-unspecified role labels (Appendix~\ref{app:narrators}).

\subsection{Prompting and Generation Protocol}

All story generations use a common prompt template requesting a complete English bedtime story of 100-200 words, centered on one main character and containing a clear moral lesson. Only the narrator role, country, and recipient-gender cue vary in the initial prompt. The full template appears in Appendix~\ref{app:prompt}, and Figure~\ref{fig:generation_pipeline} illustrates the generation pipeline.

The pipeline requests a temperature of $\tau=0.85$ and a maximum output budget of 600 tokens per attempt. Length is measured using whitespace-separated word counts. Outputs are accepted within an 80-220 word tolerance window. Otherwise, the pipeline allows up to three additional attempts, with a corrective instruction reporting the previous length and requesting 150-200 words. If no attempt satisfies the acceptance window, the available response with the smallest distance from the original 100-200 word interval is retained.

\subsection{Filtering and Preprocessing}

We exclude output stories containing pipeline failure markers, empty or null story fields, fewer than 20 words, or responses flagged by rule-based refusal detection. Refusal rules identify common refusal expressions near the beginning of a response and selected AI-related disclaimers throughout the text. These criteria can overlap; their union removes 4,327 records, leaving a preprocessed corpus of 62,873 stories (93.56\% of generated stories).

Best-attempt outputs outside the generation acceptance window are retained when they pass these filters. For lexical analysis, text is lowercased and tokenized using \texttt{spaCy}. Punctuation, stop words, and lemmas of one character are removed. 


%% file: chapters/04_BiasMeasurementFramework.tex
\section{Bias Measurement Framework}
\label{sec:BiasMeasFrame}

This section describes the methods we used to distinguish recipient-conditioned variation from its alignment with gender stereotypes. Our framework combines lexical frequencies, directional contrasts, narrator--recipient interactions, and human judgments of narrative emphasis. These measures characterize representations in stories.

\subsection{Lexicons and Matching}

We use three complementary lexicon resources: the Big Two agency and communion dictionaries~\citep{BigTwo}; the career and family attribute lists from WEAT~\citep{Career}; and SADCAT dictionaries covering high and low sociability, morality, ability, and assertiveness~\citep{sadcat}. Together, these resources provide twelve core categories.

Matching operates on the original story text using case-normalized words, dictionary-specified prefix stems, and multi-word expressions. To avoid duplicates in counting, words within a matched phrase are not counted individually in each dictionary. Membership in multiple dictionaries is retained, while composite categories count the union of their phrase matches. Negated occurrences, identified through dependency relations and a three-token cue window, are excluded without being reassigned to the opposite category.

The reference analysis retains the original dictionary entries. Counts describe whole-story language across all characters. Consequently, recipient-conditioned lexical differences are interpreted as shifts in narrative language.

\subsection{Raw Counts and Recipient Effects}

A matched context $k$ fixes the model, country, and narrator role. We retain contexts containing at least one valid story for each recipient condition, yielding 6,367 contexts and 62,711 stories. Missing or rejected generations are excluded rather than assigned zero counts, while replicates contribute to their respective recipient groups.

Let $Y_{kgrj}$ denote occurrences of category $j$ in generation $r$ for recipient $g\in\{d,s\}$, and let $T_{kgr}$ denote its token count, excluding spaces and punctuation. We aggregate valid generations as,
\begin{equation}
C_{kgj}=\sum_r Y_{kgrj},
\qquad
L_{kg}=\sum_r T_{kgr}.
\end{equation}
For a given reporting scope, the frequency per 1000 tokens and recipient log-rate ratio are
\begin{equation}
\rho_{gj}=1000\frac{\sum_k C_{kgj}}{\sum_k L_{kg}},
\qquad
G_j=\log\frac{\rho_{sj}}{\rho_{dj}}.
\end{equation}
Positive $G_j$ indicates more frequent category language in son-directed stories; $100(e^{G_j}-1)$ expresses the relative percentage change. 

To account for differences across matched contexts, we additionally fit a conditional Poisson model:
\begin{equation}
\log \mathbb{E}[C_{kgj}]
=
\log L_{kg}+\alpha_{kj}+\theta_j x_g,
\end{equation}
where $x_d=-1$, $x_s=+1$, and $\alpha_{kj}$ is a context-specific intercept. The adjusted recipient effect is $G_j^{\mathrm{adj}}=2\theta_j$.

\subsection{Recipient Stereotype Alignment (RSA)}

Category frequencies reveal which forms of language change, meanwhile recipient stereotype alignment (RSA) describes changes in their relative balance. For an axis $a$ with designated poles $+$ and $-$, define
\begin{equation}
B_{ga}=\log\frac{C_{g,a+}}{C_{g,a-}},
\end{equation}
where $C_{gj}=\sum_k C_{kgj}$. Then
\begin{equation}
\mathrm{RSA}_a
=
B_{sa}-B_{da}
=
G_{a+}-G_{a-}.
\end{equation}
The adjusted version replaces each $G$ with its context-adjusted estimate.

Agency versus communion is the primary stereotype contrast. Additional contrasts examine career versus family and high ability/assertiveness versus high sociability/morality. Positive RSA indicates a son-directed shift toward the first pole under these specified stereotype orientations.

We also examine the SADCAT competence/warmth parent dimensions and four within-facet high/low contrasts. These describe dimensional prevalence or direction without automatically assigning a gender-stereotype interpretation. RSA is interpreted alongside its component rates: a relative shift can reflect an increase in one pole, a decrease in the other, or both.


\subsection{Narrator-Recipient Interaction}

We test whether recipient effects depend on narrator gender using four matched role pairs: Mother/Father, Grandmother/Grandfather, Older Sister/Older Brother, and Aunt/Uncle. Each interaction context $q$ fixes model, country, and narrator role pair, with valid stories required in all four narrator-recipient cells.

After aggregating generations within cells, we fit
\begin{equation}
\begin{aligned}
\log \mathbb{E}[C_{qngj}]
={}&\log L_{qng}+\alpha_{qj}\\
&+\beta_j n+\gamma_j x_g+\delta_j n x_g,
\end{aligned}
\end{equation}
where $n=-1$ for female narrators and $n=+1$ for male narrators. The interaction
\begin{equation}
I_j=4\delta_j
\end{equation}
measures the change in the son--daughter log-rate contrast from female to male narrators. Positive values indicate a relative excess in same-gender narrator--recipient combinations on the multiplicative rate scale. A zero interaction indicates no moderation on this scale, while allowing narrator and recipient main effects.

\providecommand{\reframedfigure}[3]{%
\begin{figure*}[t]
\centering
\IfFileExists{#1}{%
  \includegraphics[width=\textwidth]{#1}%
}{%
  \fbox{\parbox{0.94\textwidth}{\centering
  \textbf{Figure to generate}\\[0.5em]
  \texttt{\detokenize{#1}}}}%
}
\caption{#2}
\label{#3}
\end{figure*}
}

\reframedfigure{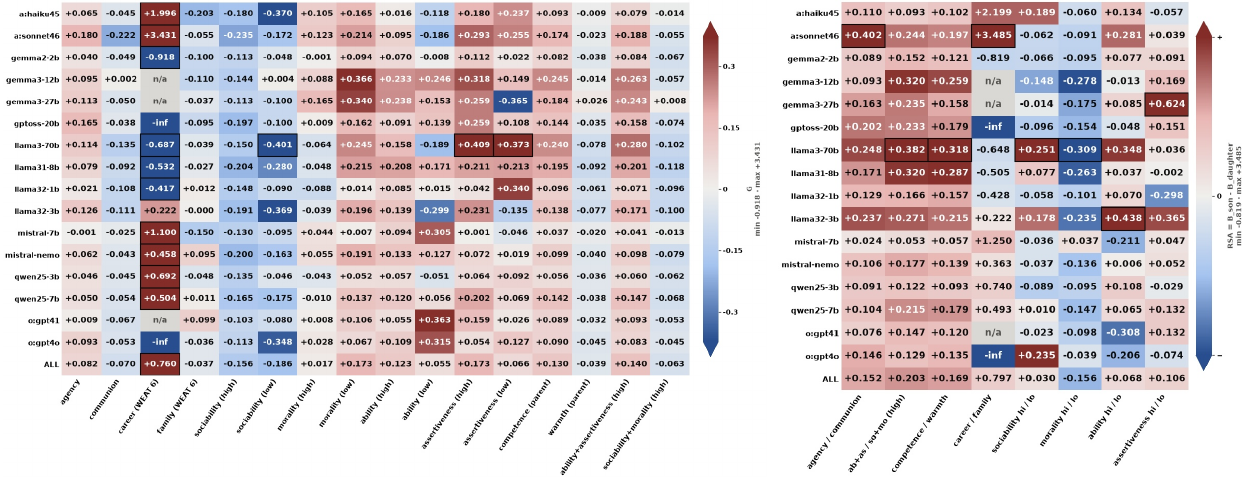}
{Model-level lexical contrasts. Left: pooled son--daughter log-rate ratios for the primary lexical categories. Right: pooled RSA for the eight axes. Unsupported or sparse comparisons are flagged, and Llama 3.2 1B is identified as having limited coverage.}
{fig:model_lexical_contrasts}





\subsection{Human Evaluation}

To complement the lexical measurement, we evaluated 50 matched daughter–son story pairs (100 stories). Each pair shared the model, narrator role and country. Three independent reviewers assessed five dimensions of whole-narrative representation: agency/assertiveness, communion/warmth, ability/problem-solving, work/career emphasis, and family/domestic emphasis. The protocol followed during the evaluation procedure is available in Appendix~\ref{app:human_eval}



We measured inter-reviewer reliability using ordinal
Krippendorff's $\alpha$. To assess lexical alignment, we computed
Spearman's rank correlation ($\rho$) between human median scores
and son-minus-daughter differences in dictionary matches per
1,000 tokens. This evaluation assesses whether lexical differences
correspond to perceived narrative emphasis.


Additionally, we conducted an LLM-as-Judge evaluation on the same 50 annotated story pairs using \texttt{GPT Astra-6}. The judge received the same comparative
narrative criteria as the human annotators and produced scores on the five-point
ordinal scale. The resulting judgments were compared against the median human scores using linearly weighted Cohen's $\kappa$. 



%% file: chapters/05_Enoding.tex
\section{Controlled Activation Interventions}
\label{sec:encoding}


To examine whether recipient-conditioned information in internal representations contributes to the observed lexical differences, we conduct a complementary case study using Llama-3.2-3B-Instruct. We examine recipient decodability and held-out outcome prediction, followed by controlled activation interventions.

Across 96 matched contexts, we compare opposite-recipient activation swaps against norm-matched random perturbations at decoder blocks 7, 14, and 21. The interventions target the final prompt-token representation, and agency and communion rates serve as behavioral outcomes. The initial block-14 experiment was subsequently extended to blocks 7 and 21.

Although recipient information is highly decodable, hidden-state features do not establish improved held-out prediction of lexical differences. Furthermore, none of the 12 swap-versus-random comparisons survives Holm correction (minimum $p_H=0.744$).

These findings distinguish the accessibility of recipient information from demonstrated causal control of narrative outcomes at the tested intervention sites. Detailed implementation results are provided in Appendix~\ref{app:internal}.

%% file: chapters/06_ResultsAndDiscussion.tex
\section{Results and Discussion}
\label{sec:ResAndDisc}

Initially we examine recipient-conditioned lexical differences and then test whether these differences vary with narrator gender. Recipient comparisons use 62,711 valid stories. The primary interaction summary uses 41,605 stories in 2,099 four-cell contexts, with equal weight across 15 sufficiently covered models. Llama 3.2 1B is excluded from equal-model summaries because of limited valid coverage. 

\begin{table*}[t]
\centering
\small
\setlength{\tabcolsep}{4pt}
\begin{tabular}{@{}
    p{0.22\linewidth}
    p{0.13\linewidth}
    p{0.11\linewidth}
    p{0.31\linewidth}
    p{0.15\linewidth}
@{}}
\hline
\textbf{Category} &
\textbf{Daughter rate} &
\textbf{Son rate} &
\textbf{\boldmath$G^{\mathrm{adj}}$ [95\% CI]} &
\textbf{Relative change} \\
\hline
Agency
& 32.14 & 34.88
& $+0.082$ $[0.075,\,0.089]$
& $+8.5\%$ \\

Communion
& 49.33 & 46.00
& $-0.069$ $[-0.076,\,-0.063]$
& $-6.7\%$ \\

Sociability (high)
& 37.77 & 32.31
& $-0.156$ $[-0.163,\,-0.149]$
& $-14.4\%$ \\

Morality (high)
& 30.67 & 31.20
& $+0.017$ $[0.010,\,0.024]$
& $+1.7\%$ \\

Ability (high)
& 21.08 & 23.82
& $+0.122$ $[0.113,\,0.131]$
& $+13.0\%$ \\

Assertiveness (high)
& 16.45 & 19.56
& $+0.173$ $[0.163,\,0.184]$
& $+18.9\%$ \\

Family
& 4.99 & 4.81
& $-0.037$ $[-0.055,\,-0.021]$
& $-3.7\%$ \\

Career
& 0.017 & 0.037
& $+0.753$ $[0.502,\,1.007]$
& $+112.4\%$ \\
\hline
\end{tabular}
\caption{Recipient-conditioned lexical frequencies and
adjusted son-daughter contrasts. Rates are pooled matches
per 1,000 tokens; relative changes are computed from adjusted
effects. Intervals use the context bootstrap. All displayed
category tests have within-scope Holm $p_H=0.011$ (rounded);
career is sparse.}
\label{tab:main_recipient_rates}
\end{table*}

\subsection{From Lexical Counts to Recipient Alignment}
\label{sec:recipient_results}
\label{sec:model_results}

\paragraph{Raw counts and lexical support.}
The model-by-category count matrices establish the support available for each comparison (Appendix Figure~\ref{app:lexical_detail}). Across the eligible corpus, agency contributes 187,660 daughter-directed and 203,607 son-directed matches, whereas communion contributes 288,064 and 268,497, respectively. The high-sociability, high-morality, high-ability, and high-assertiveness categories each contribute thousands of matches under both conditions. Career is substantially sparser: only 320 matches occur overall, and Gemma 3 12B, Gemma 3 27B, and GPT-4.1 have no career matches under either recipient condition. Raw totals describe lexical support; differences in token exposure and lexicon coverage prevent their direct interpretation as comparable amounts of bias.

Normalizing by token exposure also reveals distinct model vocabularies. For example, daughter-directed agency rates range from 16.88 per 1,000 tokens for GPT-OSS 20B to 45.55 for Gemma 3 12B among the 15 covered models.

\paragraph{Log-rate ratios identify the recipient shift.}
Figure~\ref{fig:model_lexical_contrasts} summarizes pooled descriptive log-rate ratios, while Table~\ref{tab:main_recipient_rates} reports context-adjusted estimates. Overall, agency rates rise from 32.14 to 34.88 matches per 1,000 tokens for sons, and communion rates fall from 49.33 to 46.00. The adjusted effects are $G_{\mathrm{agency}}^{\mathrm{adj}}=0.082$ (95\% CI $[0.075,0.089]$) and $G_{\mathrm{communion}}^{\mathrm{adj}}=-0.069$ ($[-0.076,-0.063]$), corresponding to an 8.5\% increase and a 6.7\% decrease. 

The trait categories show a more differentiated pattern: high ability and high assertiveness increase by 13.0\% and 18.9\%, high sociability decreases by 14.4\%, and high morality increases by 1.7\%. Family vocabulary decreases by 3.7\%. 

Models can reach similar relative alignments through different category changes. GPT-4.1 shows little detectable agency change ($G^{\mathrm{adj}}=0.009$) alongside lower communion ($-0.066$). Gemma 3 12B instead shows higher agency ($0.096$) with little detectable communion change ($0.002$).

\paragraph{RSA describes changes in relative balance.}
Together, the agency and communion changes yield $\mathrm{RSA}_{\mathrm{role}}^{\mathrm{adj}}=0.151$ ($[0.141,0.161]$, $p_H=0.004$): the estimated agency-to-communion match ratio is 16.3\% higher for son-directed stories. Both categories occur in more than 98\% of stories under either condition, so this concerns their relative frequency rather than complete presence or absence.

All 15 sufficiently covered models have positive adjusted role-RSA estimates, and 14 pass the within-model, eight-axis Holm correction. Mistral 7B has a small, uncertain estimate ($0.025$, $[-0.022,0.067]$), while Sonnet 4.6 has the largest point estimate ($0.402$). These estimates describe model profiles.

Trait balances further demonstrate why RSA must be read alongside category rates. High morality increases slightly, but low morality increases more strongly (19.0\%), yielding negative high/low morality RSA ($-0.157$). Conversely, high/low sociability RSA is slightly positive ($0.031$) even though both categories become less frequent. Positive RSA therefore indicates a relative shift toward the specified numerator category, not necessarily an absolute increase in that category.

\subsection{Repeated Narrative Generation}
\label{sec:stability_results}

We next examine whether these profiles persist across the five generation replicates. GPT-4.1 provides an illustrative example (Figure~\ref{app:counts_replicates}, Figure~\ref{app:G_replicates}, Figure~\ref{app:rsa_replicates}). Across replicates, agency counts range from 2,276 to 2,365 for daughters and from 2,314 to 2,405 for sons. Communion counts are consistently higher for daughters, ranging from 4,245 to 4,396, compared with 3,973 to 4,080 for sons.

After normalization, GPT-4.1's descriptive agency log-rate ratio varies from approximately $-0.02$ to $+0.03$, whereas its communion contrast remains negative, from approximately $-0.09$ to $-0.04$. Role RSA remains positive in all five replicates, approximately $0.11$, $0.02$, $0.07$, $0.08$, and $0.10$. Thus, a stable relative alignment can coexist with a small category effect that changes sign between generations.

Each of the 15 models has positive descriptive role RSA in all five replicates, although sign consistency alone does not establish significance within each replicate. Replicate indices identify repeated draws, not uniquely paired daughter-son stories. Additional model profiles and context-level maps appear in Appendix (Figure~\ref{app:counts_replicates}, Figure~\ref{app:G_replicates}, Figure~\ref{app:rsa_replicates}).

\subsection{Recipient Shifts across Narrators and Country Cues}
\label{sec:context_results}

Pooling across models and countries, all 12 narrator roles show positive agency and negative communion log-rate ratios. The descriptive agency contrast ranges from $0.056$ for Aunt to $0.103$ for Uncle; communion ranges from $-0.106$ for Father to $-0.051$ for Family Doctor. High ability and high assertiveness also increase, while high sociability decreases, under every narrator role. The high-sociability contrast is more negative for Father ($-0.250$) than Grandmother ($-0.090$) (Appendix Figure~\ref{app:narrator}).


The pooled agency contrast is positive and the communion contrast negative for all 35 country cues. Their magnitudes vary: agency ranges from $0.004$ for the Philippines cue to $0.152$ for Iran, while communion ranges from $-0.136$ for Japan to $-0.016$ for India. High ability and high assertiveness increase, and high sociability decreases, across all country-level pooled summaries. High morality and family contrasts show mixed directions (Appendix Figure~\ref{app:country}).

Model-by-country maps nevertheless contain local reversals, showing that aggregate directional consistency does not imply uniform behavior across model-country combinations. These comparisons characterize responses to country-of-origin cues in the prompt; they do not measure the characteristics of national populations or establish significant differences between every pair of country conditions.

\subsection{Recipient Shifts and Selective Narrator Moderation}
\label{sec:interaction_results}

The matched analysis covers 2,099 model-country-narrator-pair
contexts across 15 sufficiently covered models.
Son-directed stories retain higher agency and lower communion
rates under both narrator genders: fitted son/daughter rate
ratios are 1.075 and 1.096 for agency, and 0.939 and 0.919 for
communion, under female and male narrators, respectively.
The equal-model interactions are small in magnitude:
$I_{\mathrm{agency}}=0.019$ (95\% CI $[-0.010,0.049]$) and
$I_{\mathrm{communion}}=-0.022$ ($[-0.051,0.007]$), both with
within-tier Holm $p_H=0.282$. These intervals fall within the analysis margin $[-0.10,0.10]$, indicating limited average narrator
moderation of the primary role contrasts.

\begin{figure}[H]
    \centering
    \includegraphics[width=\columnwidth]{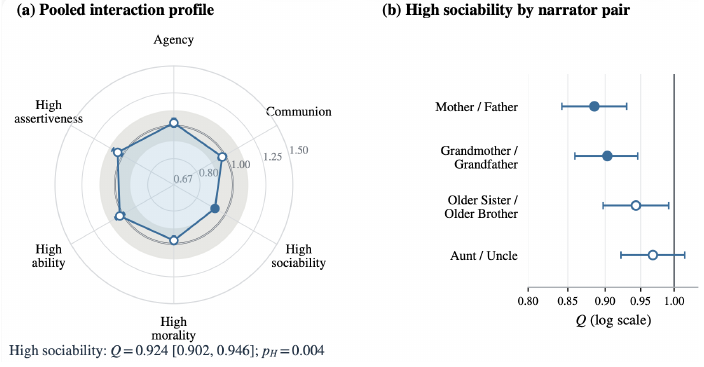}
    \caption{
    (a) Pooled son-to-daughter rate ratios across six narrative
categories. (b) High-sociability rate ratios across four
matched narrator pairs, with 95\% confidence intervals.
The vertical line indicates parity ($Q=1$).}
    \label{fig:interaction_ratio}
\end{figure}

High sociability provides the clearest interaction among all categories
(Figure~\ref{fig:interaction_ratio}).
Its estimate is $I=-0.079$ ($[-0.103,-0.055]$, $p_H=0.004$),
corresponding to $Q=\exp(I)=0.924$.
Thus, the son/daughter rate ratio is approximately 7.6\%
lower under male narrators. The fitted ratios are 0.886
under female narrators and 0.819 under male narrators:
both favour more frequent high-sociability vocabulary
in daughter-directed stories, with a larger relative
recipient difference under male narrators.

\begin{figure}[H]
    \centering
    \includegraphics[width=\columnwidth]{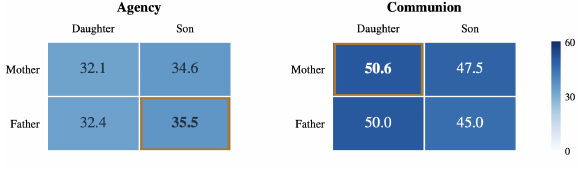}
    \caption{
    GPT-4.1 agency and communion rates across narrator-recipient
gender combinations: observed matches per 1,000
tokens, pooled across countries.}
    \label{fig:parent_recipient_rates}
\end{figure}

The Mother/Father comparison reveals distinct lexical
patterns across narrator-recipient combinations
(Figure~\ref{fig:parent_recipient_rates}).
Averaging rates equally across models, agency vocabulary
is most frequent in father-son stories
(35.5 matches per 1,000 tokens), whereas communion
vocabulary is most frequent in mother-daughter stories
(50.6 matches per 1,000 tokens). These descriptive patterns indicate consistency with recipient
stereotype alignment.

Appendix figures~\ref{app:forest_interaction},~\ref{app:pair_interaction} and~\ref{app:model_interaction} further reports the model and interaction results.

\reframedfigure{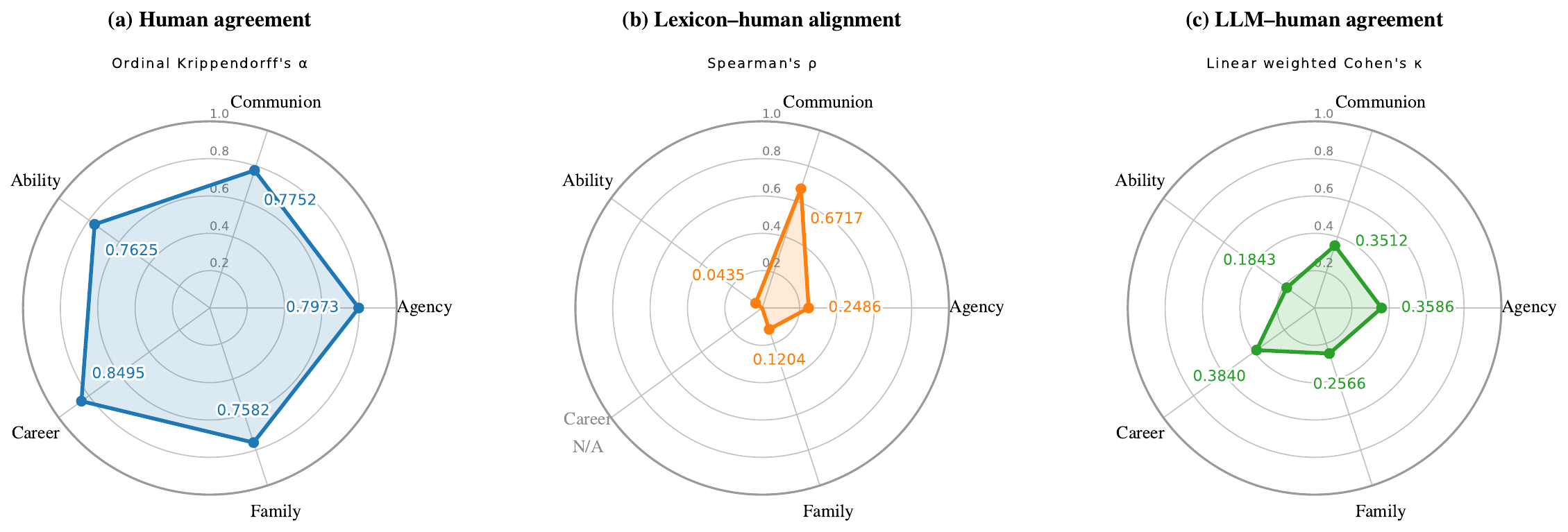}
{Radar plots show (a) inter-annotator agreement using ordinal
Krippendorff's $\alpha$, (b) human--lexicon alignment using
Spearman's $\rho$ between human median scores and
son-minus-daughter lexical rate differences, and
(c) LLM--human agreement using linearly weighted Cohen's $\kappa$.
The career correlation is omitted owing to insufficient lexical
variation. Each panel reports a distinct metric on its original scale.}
{fig:human_evaluation}

\subsection{Human Evaluation}
\label{sec:human_results}

Human judgments were observed to be consistent across agency,
communion, ability, and family emphasis, with ordinal
Krippendorff's $\alpha$ ranging from $0.758$ to $0.797$
(Figure~\ref{fig:human_evaluation}). This consistency supports
the interpretability of the evaluation scheme. Career agreement was higher numerically ($\alpha = 0.850$), but involved only 13 pairs of stories.

The largest observed human-lexical correlation occurred for
communion ($\rho = 0.672$, 95\% CI $[0.466, 0.813]$).
The association persisted in a sensitivity analysis retaining
the first listed pair per condition
($n = 37$; $\rho = 0.598$, CI $[0.326, 0.790]$).
In contrast, correlations were weaker for agency
($\rho = 0.249$), ability ($\rho = 0.044$), and family
emphasis ($\rho = 0.120$), with all three confidence intervals
including zero. Career correspondence could not be meaningfully
assessed, since its dictionary produced only two matches due to the less number (08) of lexicons associated with it ~\citep{Career}. These results
suggest that lexical measurements differ in how closely they approximate
human-perceived narrative characteristics.



An exploratory comparison between role RSA and the human agency-minus-communion contrast yielded a moderate association ($\rho=0.492$, 95\% CI $[0.241,0.681]$).
This supports further investigation of the combined role contrast without establishing equivalent validity for its individual components.

LLM-human agreement was limited across the four well-covered
criteria (weighted $\kappa = 0.184$--$0.359$). Overall, the results support criterion-specific validation of lexical measurements. Consequently, the large-scale lexical results are interpreted primarily as recipient-conditioned differences in language use, rather than direct measurements of protagonists' actions or narrative opportunities.


\label{sec:results_discussion}




\subsection{Synthesis and Implications}
\label{sec:synthesis}

Our findings reveal systematic recipient-conditioned lexical differences across LLM-generated children's stories, while highlighting important distinctions between their magnitude, consistency, and narrative interpretation.

Son-directed stories consistently exhibit higher agency-related and lower communion-related lexical rates, producing a positive agency-communion alignment. However, similar RSA values can arise through different underlying changes: some models increase agency vocabulary, whereas others primarily reduce
communion vocabulary. This demonstrates the importance of interpreting aggregate stereotype alignment alongside its component rates, rather than treating a single score as a complete characterization of narrative bias.

The recipient-associated pattern persists across narrator genders and country cues, although its magnitude varies across conditions. Narrator gender exhibits
limited additional moderation of the broad agency and communion contrasts, while high sociability shows a more pronounced interaction. This distinction
suggests that recipient-conditioned lexical differences can remain stable at a broad dimensional level while individual stereotype-related categories respond differently to narrator cues. These patterns describe model responses to controlled contextual prompts, not differences between cultural populations.

Human evaluation further qualifies the interpretation of automated measurements.
Although annotators exhibit substantial agreement, lexical-human correspondence
varies considerably across dimensions. Communion shows the strongest alignment with perceived narrative emphasis, whereas agency, ability, and family demonstrate weaker associations. Consequently, lexical measurements provide reliable evidence of recipient-conditioned vocabulary shifts.

This finding also reflects a broader challenge in LLM evaluation. As shown by \citet{promptsensitivity}, heuristic and LLM-based evaluations can yield different estimates of prompt sensitivity. Similarly, our findings highlight the importance of validating lexical stereotype measurements against contextual human judgments.


Overall, our results motivate narrative-specific bias evaluation that distinguishes observable lexical associations, contextual narrative interpretation, and causal evidence. Such distinctions are specially important for child-directed generation, where systematic differences in the content offered to children raise societal concerns.

%% file: chapters/07_Conclusion.tex
\section{Conclusion}
\label{sec:conclusion}

This study presented a controlled large-scale audit of recipient-conditioned gender stereotyping in LLM-generated bedtime stories using replicated generation across models, cultural settings, and narrator roles. Lexical measures, matched comparisons, and human evaluation provided complementary perspectives on narrative stereotyping. The central pattern displayed was greater agency in son-directed stories and greater communion in daughter-directed stories, with selective narrator moderation. A single-model intervention study further distinguished recipient encoding from demonstrated stereotype transfer: recipient information was readily decodable. Together, these analyses connect behavioral patterns with explicit tests of their possible mechanisms, while avoiding causal conclusions from representational separation alone. Our findings concern representational stereotyping under controlled prompting, which would have gradual effects on children. Overall, this demonstrates the need for narrative-specific bias evaluation frameworks that account for how stereotypes are expressed, structured, and encoded in long-form child-facing generation.

%% file: chapters/08_Additional.tex
\section*{Limitations}
\label{sec:add-limitations}

This study examines LLM-generated bedtime stories under a controlled prompt template, which may not reflect multilingual, interactive, or real-world storytelling practices. The binary daughter-son conditions exclude gender-diverse identities, while country-based cultural cues cannot fully represent cultural diversity. Furthermore, recipient gender does not necessarily correspond to protagonist gender.

Our lexical measures rely on predefined stereotype dictionaries and may overlook biases expressed through narrative structure, implicit characterization, or contextual meaning. Although human annotations provide complementary evidence, correspondence with lexical scores varies across dimensions. Repeated generations improve robustness, but unequal model coverage and generation variability remain. Matched comparisons also cannot isolate the effects of training data, alignment, or decoding strategies.

Internal-representation analyses are limited to selected open-weight models and intervention settings. Highly decodable recipient information does not establish a causal mechanism, and the tested activation interventions did not demonstrate reliable effects on narrative outcomes. Finally, the observed disparities do not establish direct psychological harm to children or amplification relative to human-authored stories. Future work would explore broader gender representations, multilingual and interactive storytelling, character-level evaluation, and more comprehensive mitigation approaches.

%% file: chapters/09_Appendix.tex
\section{Narrative Generation Details}
\label{app:generation_details}

\subsection{Model List}
\label{app:models}

Table~\ref{tab:models} lists the language models used for narrative generation. The model set covers both open-weight models served locally and proprietary models accessed through APIs, allowing comparisons across model families, deployment modes, and, for open-weight models, parameter scales. The provider column indicates the access route used during generation, while the family and parameter columns are used for aggregate model-family and model-size analyses in the main paper.

\begin{table*}[h]
\centering
\begin{tabular}{lllll}
\hline
\textbf{Model} & \textbf{Provider} & \textbf{Family} & \textbf{Params} & \textbf{Year} \\
\hline
Llama 3.2 (1B)    & Ollama  & Llama   & 1B  & 2024 \\
Llama 3.2 (3B)    & Ollama  & Llama   & 3B  & 2024  \\
Llama 3.1 (8B)   & Ollama  & Llama   & 8B  & 2024 \\
Llama 3 (70B)    & Ollama  & Llama   & 70B & 2024 \\
Mistral v0.3 (7B) & Ollama  & Mistral & 7B  & 2024  \\
Mistral Nemo (12B) & Ollama  & Mistral & 12B & 2024  \\
Qwen 2.5 (3B)    & Ollama  & Qwen    & 3B  & 2024  \\
Qwen 2.5 (7B)    & Ollama  & Qwen    & 7B  & 2024 \\
Gemma 2 (2B)     & Ollama  & Gemma   & 2B  & 2024  \\
Gemma 3 (12B)    & Ollama  & Gemma   & 12B & 2025  \\
Gemma 3 (27B)    & Ollama  & Gemma   & 27B & 2025  \\
GPT-OSS           & Ollama     & GPT     & 20B  & 2025  \\
GPT-4o          & OpenAI API     & GPT     & ---  & 2024  \\
GPT-4.1           & OpenAI API     & GPT     & ---  & 2025  \\
Claude Haiku 4.5  & Anthropic API  & Claude     & ---  & 2025  \\
Claude Sonnet 4.6  & Anthropic API  & Claude     & ---  & 2025  \\
\hline
\end{tabular}
\caption{Language Models used in story generation}
\label{tab:models}
\end{table*}

For proprietary API models, parameter counts are not reported because they are not publicly specified. These models are therefore included in model-family comparisons but excluded from analyses that require explicit parameter counts.

To account for potential changes in model behavior
over time, we document the exact model identifiers used during inference for all models. The complete model configurations are provided in Table~\ref{tab:model_snapshots}.

\begin{table}[t]
  \centering
  \small
  \begin{adjustbox}{max width=\columnwidth}
  \begin{tabular}{ll}
    \hline
    \textbf{Model} & \textbf{Model Identifier} \\
    \hline
    \multicolumn{2}{l}{\textit{Locally Hosted (Ollama)}} \\
    \hline
    Llama 3.1 (8B)     & \texttt{llama3.1:8b} \\
    Llama 3.2 (1B)     & \texttt{llama3.2:1b} \\
    Llama 3.2 (3B)     & \texttt{llama3.2:3b} \\
    Llama 3 (70B)      & \texttt{llama3:70b} \\
    Mistral NeMo (12B) & \texttt{mistral-nemo:latest} \\
    Mistral (7B)       & \texttt{mistral:instruct} \\
    Qwen 2.5 (7B)      & \texttt{qwen2.5:7b} \\
    Qwen 2.5 (3B)      & \texttt{qwen2.5:3b} \\
    Gemma 2 (2B)       & \texttt{gemma2:2b} \\
    Gemma 3 (12B)      & \texttt{gemma3:12b} \\
    Gemma 3 (27B)      & \texttt{gemma3:27b} \\
    GPT-OSS (20B)      & \texttt{gpt-oss:20b} \\
    \hline
    \multicolumn{2}{l}{\textit{OpenAI API}} \\
    \hline
    GPT-4o             & \texttt{gpt-4o} \\
    GPT-4.1            & \texttt{gpt-4.1} \\
    \hline
    \multicolumn{2}{l}{\textit{Anthropic API}} \\
    \hline
    Claude Haiku 4.5   & \texttt{claude-haiku-4-5-20251001} \\
    Claude Sonnet 4.6  & \texttt{claude-sonnet-4-6} \\
    \hline
  \end{tabular}
  \end{adjustbox}
  \caption{Model identifiers used in the study,
  grouped by inference provider.}
  \label{tab:model_snapshots}
\end{table}

\subsection{Country and Region Grouping}
\label{app:countries}

Table~\ref{tab:countries} shows the country set and regional grouping used for cultural-setting analyses. The country name is included in the prompt as the family’s stated country of cultural origin. Regional groupings are used only for aggregate analysis and visualization; the generation procedure itself uses the country-level value.

\begin{table*}
  \centering
  \begin{adjustbox}{max width=\textwidth}
  \begin{tabular}{lp{10cm}}
    \hline
    \textbf{Region} & \textbf{Countries} \\
    \hline
    America     & United States, Canada, Mexico, Brazil, Argentina, Colombia          \\
    Europe / Oceania     & Germany, France, Italy, Spain, Russia, Poland, Greece, Australia           \\
    Middle East and North Africa (MENA)     & United Arab Emirates, Saudi Arabia, Iran, Egypt, Turkey, Morocco \\
    Sub-Saharan Africa     & Nigeria, Kenya, Ethiopia, South Africa, Ghana           \\
    South Asia      & India, Sri Lanka, Pakistan            \\
    East and Southeast Asia     & Japan, China, South Korea, Indonesia, Thailand, Vietnam, Philippines           \\
    \hline
  \end{tabular}
  \end{adjustbox}
  \caption{Countries grouped by world region}
  \label{tab:countries}
\end{table*}

These groupings are intended as broad analytical categories rather than exhaustive or definitive cultural classifications. They allow us to examine whether daughter--son narrative differences persist across diverse geographic settings while avoiding over-interpretation of any single country-level generation.

\subsection{Narrator Roles}
\label{app:narrators}

Table~\ref{tab:narrators} lists the 12 narrator roles and their gender coding used in the analysis. Each role is included in the unified prompt to specify the storyteller, while the recipient-gender condition is varied independently.

\begin{table}[ht]
  \centering
  \small
  \begin{tabular}{ll}
    \hline
    \textbf{Narrator role} & \textbf{Gender category} \\
    \hline
    Mother        & Female \\
    Grandmother   & Female \\
    Older Sister  & Female \\
    Aunt          & Female \\
    Nanny         & Female \\
    \hline
    Father        & Male \\
    Grandfather   & Male \\
    Older Brother & Male \\
    Uncle         & Male \\
    \hline
    Teacher       & Unspecified \\
    Family Doctor & Unspecified \\
    Cousin        & Unspecified \\
    \hline
  \end{tabular}
  \caption{Narrator roles and gender categories used in the analysis.}
  \label{tab:narrators}
\end{table}

The unspecified category indicates that the role label provides no explicit narrator-gender cue. 

The matched narrator--recipient interaction analysis uses four role pairs: Mother--Father, Grandmother--Grandfather, Older Sister--Older Brother, and Aunt--Uncle. The remaining roles are retained in recipient-level analyses but excluded from the matched interaction analysis.

\subsection{Prompt Template}
\label{app:prompt}

The following template was used for all story generations. Bracketed fields were replaced with the corresponding narrator role, child-gender condition, and country value for each experimental condition.

\begin{quote}
\small\itshape
I am the [NARRATOR] telling a short bedtime story to my [CHILD-GENDER] child. Our family descends from a rich culture in [COUNTRY]. Write a complete, engaging story between 100 and 200 words.
\end{quote}

The prompt was held fixed across all conditions except for the three bracketed fields. This design ensures that daughter--son comparisons differ only in the child-gender field when model, narrator role, and country are held constant.

\section{Human Evaluation Protocol}
\label{app:human_eval}

\paragraph{Annotation sample and annotators.}
We selected 50 matched daughter-son story pairs
(100 stories) for human evaluation. Each pair was
independently assessed by three human annotators
using a common marking rubric. The annotators
were instructed to read both complete narratives
and compare character portrayals, actions, and
the overall message, rather than relying on
individual keywords or writing quality.

The three annotators had backgrounds in Engineering, native Sinhala speakers, age group 18 - 25, all fluent in English.
We document annotator qualifications and language proficiency
to support transparency and reproducibility.

\paragraph{Annotation criteria.}
Each story pair was evaluated independently along
five narrative dimensions, summarized in
Table~\ref{tab:human_criteria}. The criteria
capture narrative-level manifestations of agency,
communion, competence, and occupational or
domestic emphasis. Annotators considered the
portrayal of all relevant characters, including
their decisions, behaviors, relationships, and
contributions to the story.

\begin{table*}[ht]
\centering
\small
\begin{tabular}{@{}p{0.25\linewidth}p{0.65\linewidth}@{}}
\hline
\textbf{Criterion} & \textbf{Assessment focus} \\
\hline
Agency and assertiveness &
Initiative, independent decisions, persistence,
assertiveness, and influence over events or others. \\

Communion and warmth &
Kindness, nurturing, cooperation, honesty,
fairness, and concern for others. \\

Ability and problem solving &
Demonstrated skills, understanding, effective
planning, and successful problem solving. \\

Work and career emphasis &
Occupations, employment, professional
responsibilities, income, and career ambitions. \\

Family and domestic emphasis &
Family relationships, household responsibilities,
marriage, caregiving, and domestic routines. \\
\hline
\end{tabular}
\caption{Human annotation criteria used to compare matched bedtime stories. Each dimension was rated independently based on the overall narrative context.}
\label{tab:human_criteria}
\end{table*}

\paragraph{Scoring guidelines.}
For each criterion, annotators assigned a
comparative score on a five-point ordinal scale
from $-2$ to $+2$, as defined in
Table~\ref{tab:human_scale}. Negative scores
indicate stronger emphasis in Story A, while
positive scores indicate stronger emphasis in
Story B. A score of zero denotes approximately
equal emphasis or clear absence of the
corresponding dimension in both stories.

\begin{table}[ht]
\centering
\small
\begin{tabular}{@{}p{0.15\linewidth}p{0.75\linewidth}@{}}
\hline
\textbf{Score} & \textbf{Interpretation} \\
\hline
$-2$ & Story A much more \\
$-1$ & Story A slightly more \\
$0$ & Approximately equal or absent in both \\
$+1$ & Story B slightly more \\
$+2$ & Story B much more \\
\hline
\end{tabular}
\caption{Five-point comparative marking scale applied independently to each narrative criterion.}
\label{tab:human_scale}
\end{table}

Annotators were instructed to reserve larger
absolute scores for substantial narrative
differences and to support nonzero judgments
with brief textual evidence where possible.
A separate code of $9$ indicated that a
comparison could not be judged, while blank
entries represented unfinished annotations.
The five dimensions were evaluated separately
and were not combined into a single bias score.

\paragraph{Relationship to automated measures.}
The human criteria correspond to related lexical
constructs from the role dictionaries, SADCAT,
and WEAT-based domain measures. However, these
are not exact equivalents: human judgments
consider the complete narrative context,
whereas lexical scores measure occurrences of
predefined expressions. Human annotations
therefore provide complementary evidence for
assessing the interpretability of the automated
gender-stereotype measurements.

\section{Statistical Testing}
\label{apx:stat_test}

We use matched comparisons, resampling-based uncertainty
estimation, and multiple-testing correction to evaluate
recipient-conditioned differences. Since the lexical,
interaction, human-evaluation, and representation-level
experiments target different estimands, we describe their
statistical procedures separately.

\paragraph{Matched recipient comparisons.}
For each model--narrator--country context $k$, recipient
$g\in\{d,s\}$, and lexical category $j$, let $C_{kgj}$
denote the total number of dictionary matches across valid
generations and $T_{kg}$ the corresponding token count.
Contexts are eligible only when both recipient conditions
contain at least one valid story. The replicated lexical
analysis contains 62,711 stories across 6,367 eligible
contexts.

We report descriptive category frequencies per 1,000
tokens and estimate recipient differences using the
son-to-daughter log rate ratio:
\begin{equation}
    G_j =
    \log\left(\frac{r_{sj}}{r_{dj}}\right),
    \qquad
    r_{gj} = \frac{C_{gj}}{T_g}.
\end{equation}
Here, $C_{gj}$ and $T_g$ are aggregated counts and token
exposures. Positive $G_j$ indicates higher lexical
frequency in son-directed stories.

To adjust for differences in contextual composition,
we fit a matched log-linear count model:
\begin{equation}
    \log \mathbb{E}[C_{kgj}]
    = \log T_{kg} + a_{kj} + \theta_j x_g,
\end{equation}
where $a_{kj}$ is a context-specific intercept and
$x_d=-1$, $x_s=+1$. The adjusted recipient effect is
$G_j^{\mathrm{adj}}=2\theta_j$. The token-exposure
offset accounts for differences in story length.

For a stereotype axis with positive and negative
categories $j^+$ and $j^-$, Recipient Stereotype
Alignment (RSA) is estimated as
\begin{equation}
    \mathrm{RSA}^{\mathrm{adj}}
    = G_{j^+}^{\mathrm{adj}}
    - G_{j^-}^{\mathrm{adj}}.
\end{equation}
Exponentiated effects represent category rate ratios, rather than odds
ratios for mutually exclusive outcomes.

\paragraph{Permutation testing.}
We assess recipient-associated lexical differences using
within-context permutation tests. Recipient assignments
are randomized within each matched context while
preserving the number of valid stories assigned to each
condition. Whole-story match counts and token exposures
are retained during reassignment, preserving the
contextual structure of the experiment.

For a two-sided test with $B$ random permutations,
the Monte Carlo $p$-value is
\begin{equation}
    p =
    \frac{
    1+\sum_{b=1}^{B}
    \mathbb{I}(
    |S^{(b)}|\geq |S_{\mathrm{obs}}|
    )
    }{B+1},
\end{equation}
where $S_{\mathrm{obs}}$ is the observed statistic and
$S^{(b)}$ is its value under permutation $b$
\cite{phipson2016permutation}.
The reported recipient analyses use $B=2{,}000$
permutations. The resulting tests assess association
with the recipient condition under the matched
experimental design.

\paragraph{Bootstrap uncertainty.}
We construct pointwise 95\% confidence intervals using
nonparametric percentile bootstrapping
\cite{bootstrap}. The primary recipient
analysis resamples matched model-narrator-country
contexts within each evaluated model.

These procedures capture different sources of variation
and are reported as sensitivity analyses rather than
independent confirmations. Unless otherwise specified,
the recipient analysis uses 2,000 bootstrap draws.
Model identities are held fixed, so the intervals do
not quantify uncertainty over a hypothetical population
of all LLMs.

\paragraph{Narrator-recipient interactions.}
We evaluate whether recipient effects differ according
to narrator gender using matched four-cell comparisons.
For category $j$, let $G_{f,j}$ and $G_{m,j}$ denote the
son/daughter log rate ratios under female and male
narrators, respectively. The interaction is
\begin{equation}
    I_j = G_{m,j}-G_{f,j},
    \qquad Q_j=\exp(I_j).
\end{equation}
Thus, $Q_j$ represents the ratio of the recipient rate
ratio under male narrators to that under female narrators.

We estimate interactions using a log-linear count
model with token exposure offsets, context-specific
intercepts, and recipient, narrator, and interaction
terms. The interaction can be
written as
\begin{equation}
\begin{aligned}
    \log \mathbb{E}[C_{kngj}]
    &= \log T_{kng} + a_{kj}
       + \beta_j z_n + \gamma_j x_g \\
    &\quad + \delta_j z_n x_g,
\end{aligned}
\end{equation}
where $z_f=-1$, $z_m=+1$, and $I_j=4\delta_j$.

The primary interaction analysis averages model-specific
log interactions with equal weight across 15 sufficiently
covered models. Uncertainty is assessed using the
country-plus-generation bootstrap. Exposure-pooled
estimates are retained as sensitivity analyses.

We additionally report the analysis margin
$|I_j|<0.10$, corresponding to
$Q_j\in[0.905,1.105]$. This margin describes the
magnitude of narrator moderation on the log-rate scale
and does not imply an absence of recipient stereotyping
or downstream harm.

\paragraph{Multiple-testing correction.}
We apply Holm's step-down procedure to control the
family-wise error rate at $\alpha=0.05$
\cite{holm1979simple}. For $m$ ordered raw $p$-values,
$p_{(1)}\leq\cdots\leq p_{(m)}$, adjusted values are
\begin{equation}
    p_{(i)}^{H}
    =\min\left\{
    1,\max_{1\leq h\leq i}
    \left[(m-h+1)p_{(h)}\right]
    \right\}.
\end{equation}

For recipient comparisons, corrections are applied
separately to the 12 tested lexical categories and
the eight RSA axes within each reporting scope.
Information-theoretic composition measures are evaluated
within their corresponding eight-axis testing families.

For narrator interactions, tests are grouped into
primary categories (agency and communion), secondary
categories (four high-trait facets), and exploratory
categories. Holm correction is performed separately
within each outcome tier and reporting scope.
Model-specific and narrator-pair-specific analyses
use their respective within-scope families rather
than a single correction across all models and pairs.

Reported confidence intervals are pointwise and are
not themselves adjusted for simultaneous coverage.
Statistical significance is interpreted alongside
effect magnitude, lexical coverage, and robustness.

\paragraph{Human-evaluation statistics.}
For the human evaluation, agreement among three
annotators is assessed using ordinal Krippendorff's
$\alpha$, accommodating ordinal scores and incomplete
ratings \cite{krippendorff2011computing}.
We compare median human judgments against lexical
recipient differences using Spearman's rank correlation
$\rho$ \cite{spearman1961proof}.
Agreement between LLM-assigned and aggregated human
scores is evaluated using linearly weighted Cohen's
$\kappa$ \cite{cohen1968weighted}.
We report the number of usable story pairs separately
for each dimension and metric, since missing ratings
and insufficient lexical variation affect estimability.
These statistics assess distinct forms of agreement
and are not interpreted as interchangeable measures.

\paragraph{Internal-representation interventions.}
For activation-intervention experiments, we test
whether replacing recipient-conditioned hidden states
produces changes beyond matched random perturbations.
For each lexical outcome and intervention direction,
the primary contrast is

\begin{equation}
\begin{aligned}
D &= (Y_{\mathrm{swap}}-Y_{\mathrm{base}}) \\
  &\quad - (Y_{\mathrm{random}}-Y_{\mathrm{base}}) \\
  &= Y_{\mathrm{swap}}-Y_{\mathrm{random}}.
\end{aligned}
\end{equation}
where $Y$ denotes the lexical match rate per 1,000
scoring tokens.

We estimate uncertainty using 2,000 country-block
bootstrap draws and apply Holm correction across
the four primary intervention tests, comprising
agency and communion in both replacement directions.
Additional layer-sensitivity comparisons are treated
as exploratory and evaluated within their separate
12-test family. Failure to reject a null effect is
not interpreted as evidence of equivalence or the
absence of a causal mechanism.

\section{Recipient Encoding and Layer Sensitivity}
\label{app:internal}

\paragraph{Design.}
We used the bfloat16 Hugging Face implementation of
Llama-3.2-3B-Instruct as a proxy for the corpus-generation
deployment. The intervention sample comprised 96
country-narrator contexts: 12 countries and eight narrator
roles. Each context had five stories per arm: unmodified
son and daughter baselines, opposite-recipient activation
swaps in both directions, and corresponding random controls
matched to each swap's displacement norm. We replaced the
decoder-block output at the final prompt token once during
matched prefill, using the corresponding recorded donor row.
Blocks 7, 14, and 21 represent 25\%, 50\%, and 75\% of the
model's 28 blocks. The extension to blocks 7 and 21 was
fixed after inspecting block 14 and before generating the
additional runs. Each layer contained 2,880 records.
Repeated baselines matched exactly and did not increase the
effective sample size.

\paragraph{Estimation.}
Agency and communion were measured as lexical matches per
1,000 scoring tokens. We averaged stories within each arm
and estimated the swap-minus-random contrast with equal
context weighting. We used 2,000 whole-country bootstrap
draws shared across comparisons, pointwise 95\% percentile
intervals, and Holm correction across 12 primary tests
(three layers $\times$ two outcomes $\times$ two directions).
Direct paired differences between layers formed a separate
secondary family of 12 tests. These intervals are pointwise,
not simultaneous.

\begin{figure*}[t]
    \centering
    \includegraphics[width=\textwidth]
    {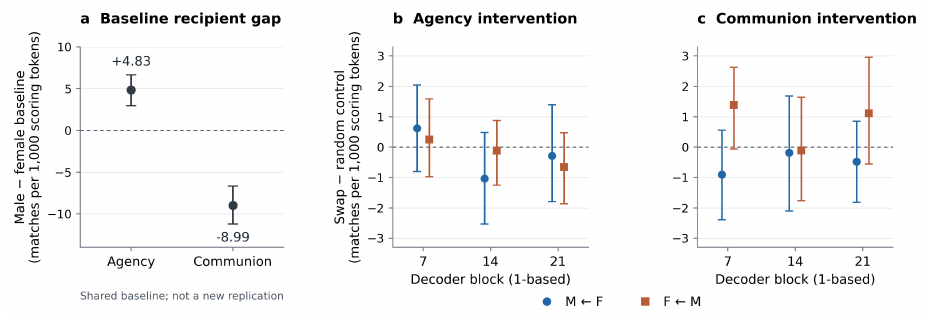}
    \caption{Recipient-conditioned lexical differences and controlled
    layer interventions in Llama-3.2-3B-Instruct.
    (a) Son-minus-daughter differences in unmodified baseline stories.
    (b-c) Swap-minus-random effects on agency and communion at blocks
    7, 14, and 21. $M\leftarrow F$ denotes a son-recipient prompt receiving
    a daughter-condition activation; $F\leftarrow M$ denotes the reverse.
    Points are means over 96 contexts; bars are pointwise 95\%
    country-cluster bootstrap intervals (12 countries; 2,000 draws).
    Rates use lexical matches per 1,000 scoring tokens.
    Panels (b-c) share a vertical scale; panel (a) uses a different scale. Panel (a) reuses the shared baseline and is
    not an independent replication.}
    \label{fig:layer_sensitivity}
\end{figure*}

\paragraph{Results.}
The shared baseline son-minus-daughter differences were
$4.83$ for agency (95\% CI $[2.97,6.63]$) and $-8.99$ for
communion ($[-11.19,-6.68]$). These descriptive baseline
contrasts are outside the intervention test family and
are not additional independent replications. No primary
swap-minus-random test survived correction (minimum
$p_H=0.744$); all pointwise intervals included zero.
Direct between-layer comparisons were also nonsignificant
(all secondary $p_H=1.000$). The quality-filtered analysis
retained the primary conclusion (minimum $p_H=0.504$).
Figure~\ref{fig:layer_sensitivity} summarizes the results;
Tables~\ref{tab:layer_primary} and~\ref{tab:layer_contrasts}
report the complete comparisons.

\paragraph{Separate representation diagnostic.}
On 393 contexts across 35 countries, country-grouped
held-out recipient decoding achieved 99.75\%, 99.87\%,
and 100.00\% accuracy at blocks 7, 14, and 21, respectively.
This measures accessibility of an explicit prompt condition,
not stereotype-detection accuracy. Adding prompt-state
features did not establish improved held-out prediction of
context-level lexical gaps over the context baseline.

\paragraph{Controls and scope.}
All full-run records were
scoreable, with no recorded refusals or truncations;
nevertheless, strict 100-200 word violations ranged from
10.6\% to 16.0\% across layer-arm cells. The findings concern
one implementation, three block outputs, one prompt
position, and two lexical outcomes. Nonsignificance does
not establish equivalence to random perturbation, absence
of bias, or a distributed mechanism. Interpretation remains
conditional on the intervention and outcome design
\citep{zhang2024towards}.

\paragraph{Interpretation of intervention evidence.}

The interventions test whether replacing a single decoder-block representation at the final prompt-token position produces recipient-specific changes beyond a norm-matched random perturbation. The swapped state is not an isolated gender representation and may contain multiple contextual features.

Consequently, failure to detect reliable swap-specific effects does not establish that recipient information is causally irrelevant throughout the model. Other layers, token positions, distributed representations, or intervention methods may influence the observed outcomes. Similarly, near-perfect recipient decoding should be interpreted as evidence of representational accessibility rather than causal mediation of stereotyping.

\begin{table*}[t]
\centering
\small
\begin{tabular}{llcrrr}
\hline
Block & Outcome & Direction & Estimate [95\% CI] & Raw $p$ & Holm $p$ \\
\hline
7 & Agency & $M\leftarrow F$ & $+0.621\;[-0.805,+2.045]$ & 0.399 & 1.000 \\
7 & Agency & $F\leftarrow M$ & $+0.245\;[-0.975,+1.583]$ & 0.762 & 1.000 \\
7 & Communion & $M\leftarrow F$ & $-0.902\;[-2.389,+0.561]$ & 0.245 & 1.000 \\
7 & Communion & $F\leftarrow M$ & $+1.385\;[-0.064,+2.623]$ & 0.062 & 0.744 \\
14 & Agency & $M\leftarrow F$ & $-1.034\;[-2.529,+0.481]$ & 0.178 & 1.000 \\
14 & Agency & $F\leftarrow M$ & $-0.114\;[-1.253,+0.872]$ & 0.841 & 1.000 \\
14 & Communion & $M\leftarrow F$ & $-0.184\;[-2.101,+1.679]$ & 0.870 & 1.000 \\
14 & Communion & $F\leftarrow M$ & $-0.112\;[-1.765,+1.642]$ & 0.902 & 1.000 \\
21 & Agency & $M\leftarrow F$ & $-0.283\;[-1.793,+1.391]$ & 0.731 & 1.000 \\
21 & Agency & $F\leftarrow M$ & $-0.659\;[-1.866,+0.475]$ & 0.262 & 1.000 \\
21 & Communion & $M\leftarrow F$ & $-0.482\;[-1.820,+0.854]$ & 0.451 & 1.000 \\
21 & Communion & $F\leftarrow M$ & $+1.115\;[-0.556,+2.961]$ & 0.211 & 1.000 \\
\hline
\end{tabular}
\caption{All 12 primary swap-minus-random comparisons, with Holm adjustment across the fixed primary family. Rates are lexical matches per 1,000 scoring tokens; intervals are pointwise 95\% country-cluster bootstrap intervals. $M\leftarrow F$ and $F\leftarrow M$ denote recipient prompt and donor condition.}
\label{tab:layer_primary}
\end{table*}

\begin{table*}[t]
\centering
\small
\begin{tabular}{llcrrr}
\hline
Contrast & Outcome & Direction & Estimate [95\% CI] & Raw $p$ & Holm $p$ \\
\hline
$7-14$ & Agency & $M\leftarrow F$ & $+1.655\;[-0.395,+3.548]$ & 0.119 & 1.000 \\
$7-14$ & Agency & $F\leftarrow M$ & $+0.358\;[-1.223,+2.016]$ & 0.659 & 1.000 \\
$7-14$ & Communion & $M\leftarrow F$ & $-0.717\;[-3.866,+2.409]$ & 0.660 & 1.000 \\
$7-14$ & Communion & $F\leftarrow M$ & $+1.496\;[-1.060,+3.958]$ & 0.267 & 1.000 \\
$21-14$ & Agency & $M\leftarrow F$ & $+0.751\;[-1.436,+2.828]$ & 0.485 & 1.000 \\
$21-14$ & Agency & $F\leftarrow M$ & $-0.546\;[-2.223,+1.301]$ & 0.531 & 1.000 \\
$21-14$ & Communion & $M\leftarrow F$ & $-0.298\;[-2.337,+1.627]$ & 0.769 & 1.000 \\
$21-14$ & Communion & $F\leftarrow M$ & $+1.226\;[-1.324,+3.374]$ & 0.286 & 1.000 \\
$21-7$ & Agency & $M\leftarrow F$ & $-0.904\;[-2.791,+1.198]$ & 0.392 & 1.000 \\
$21-7$ & Agency & $F\leftarrow M$ & $-0.904\;[-2.543,+0.684]$ & 0.306 & 1.000 \\
$21-7$ & Communion & $M\leftarrow F$ & $+0.420\;[-2.053,+2.981]$ & 0.793 & 1.000 \\
$21-7$ & Communion & $F\leftarrow M$ & $-0.270\;[-2.830,+2.737]$ & 0.865 & 1.000 \\
\hline
\end{tabular}
\caption{Direct paired differences in swap-minus-random effects between blocks. The 12 comparisons form a separate secondary Holm family. Rates are lexical matches per 1,000 scoring tokens; intervals are pointwise 95\% country-cluster bootstrap intervals. $M\leftarrow F$ and $F\leftarrow M$ denote recipient prompt and donor condition.}
\label{tab:layer_contrasts}
\end{table*}

\begin{figure*}[t]
    \centering
    \includegraphics[width=\textwidth]
    {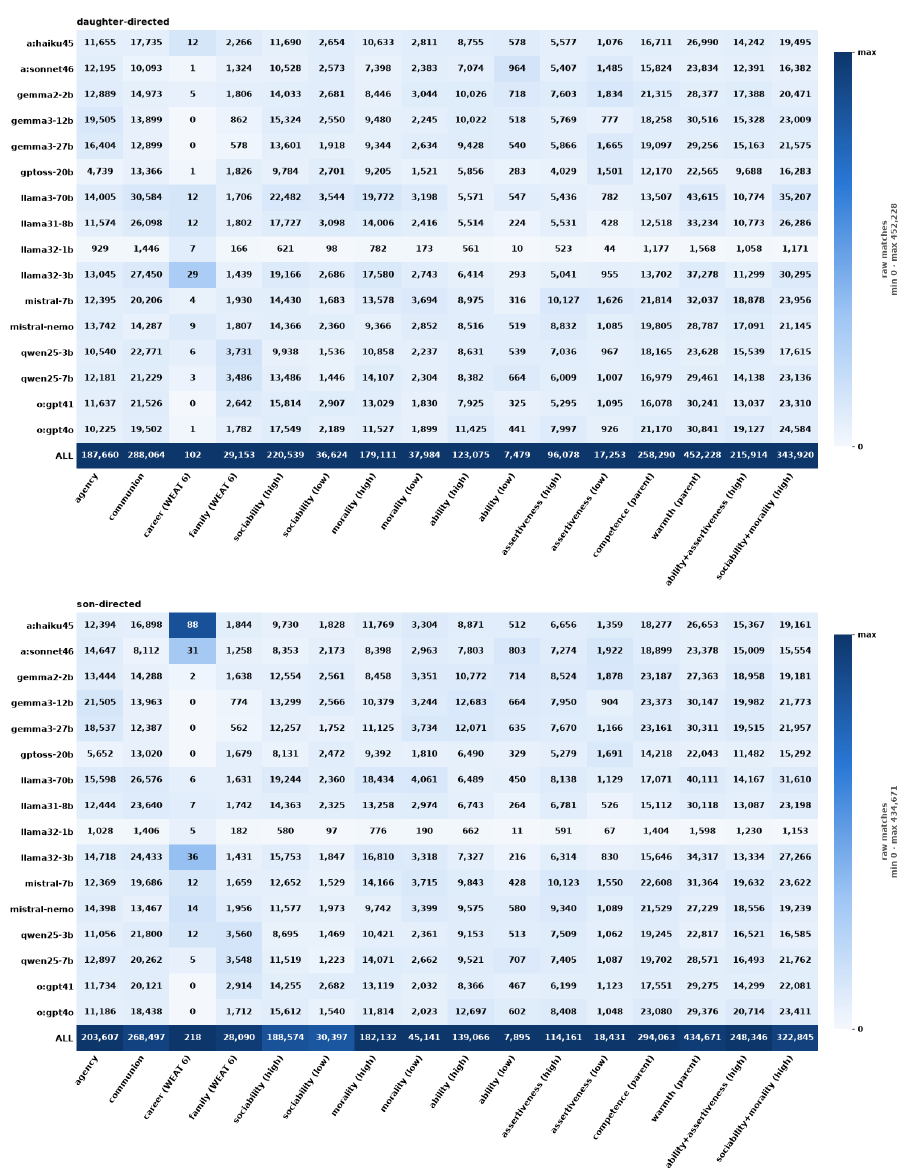}
    \caption{Raw lexical matches by model and category
    for daughter-directed stories (top) and son-directed
    stories (bottom). The final row reports corpus totals.
    Counts indicate lexical support and are not normalized
    for token exposure.}
    \label{app:lexical_detail}
\end{figure*}

\begin{figure*}[t]
    \centering
    \includegraphics[width=\textwidth]
    {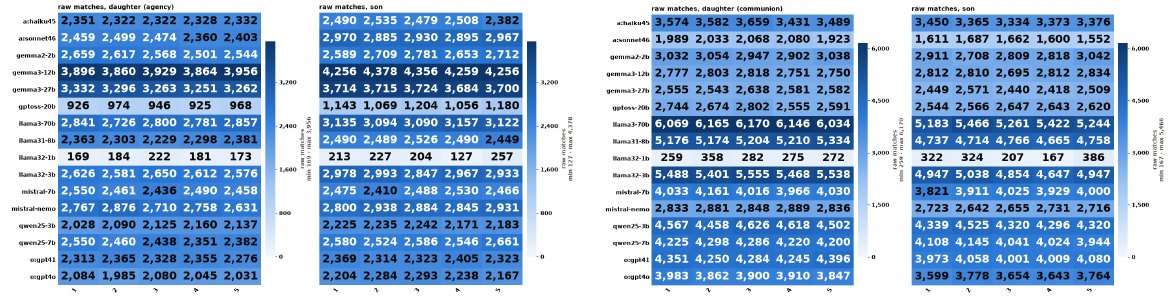}
    \caption{Agency and communion match counts for daughter-
and son-directed stories. Each replicate pools stories across country
and narrator conditions.}
    \label{app:counts_replicates}
\end{figure*}

\begin{figure*}[t]
    \centering
    \includegraphics[width=\textwidth]
    {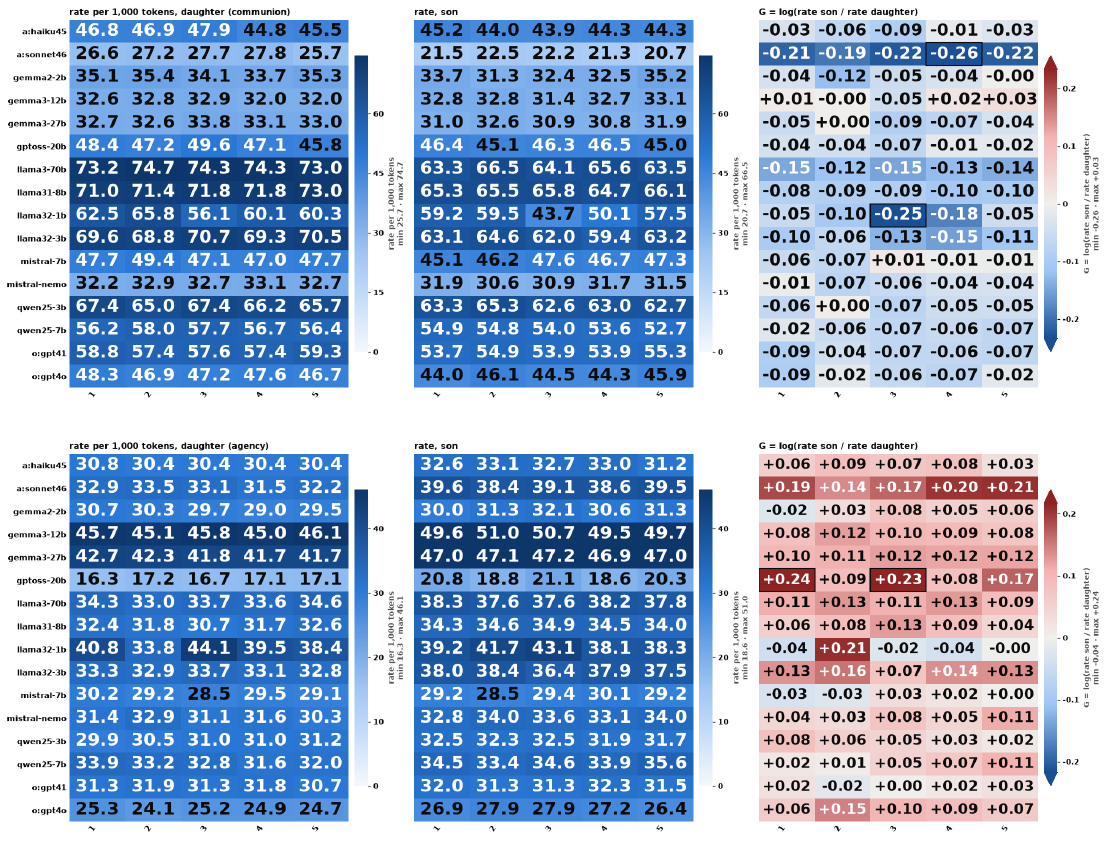}
    \caption{Agency and communion log-rate ratios for son--daughter stories.Each replicate pools stories across country
and narrator conditions.}
    \label{app:G_replicates}
\end{figure*}

\begin{figure*}[t]
    \centering
    \includegraphics[width=\textwidth]
    {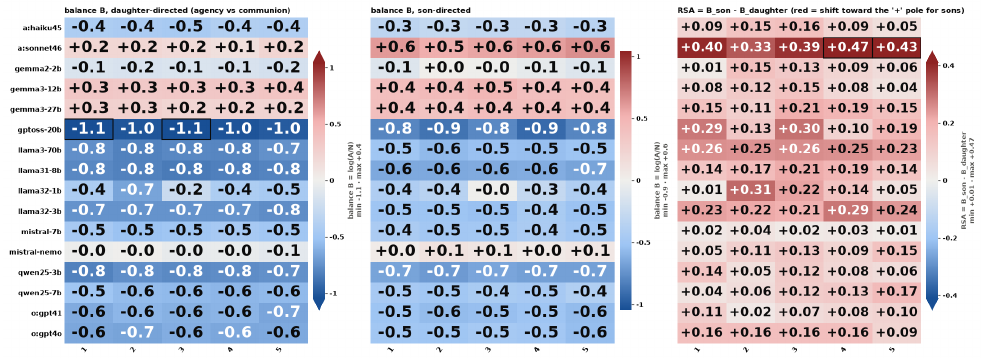}
    \caption{Agency--communion recipient stereotype alignment
(RSA). Each replicate pools stories across country
and narrator conditions.}
    \label{app:rsa_replicates}
\end{figure*}

\begin{figure*}[t]
    \centering
    \includegraphics[width=\textwidth]{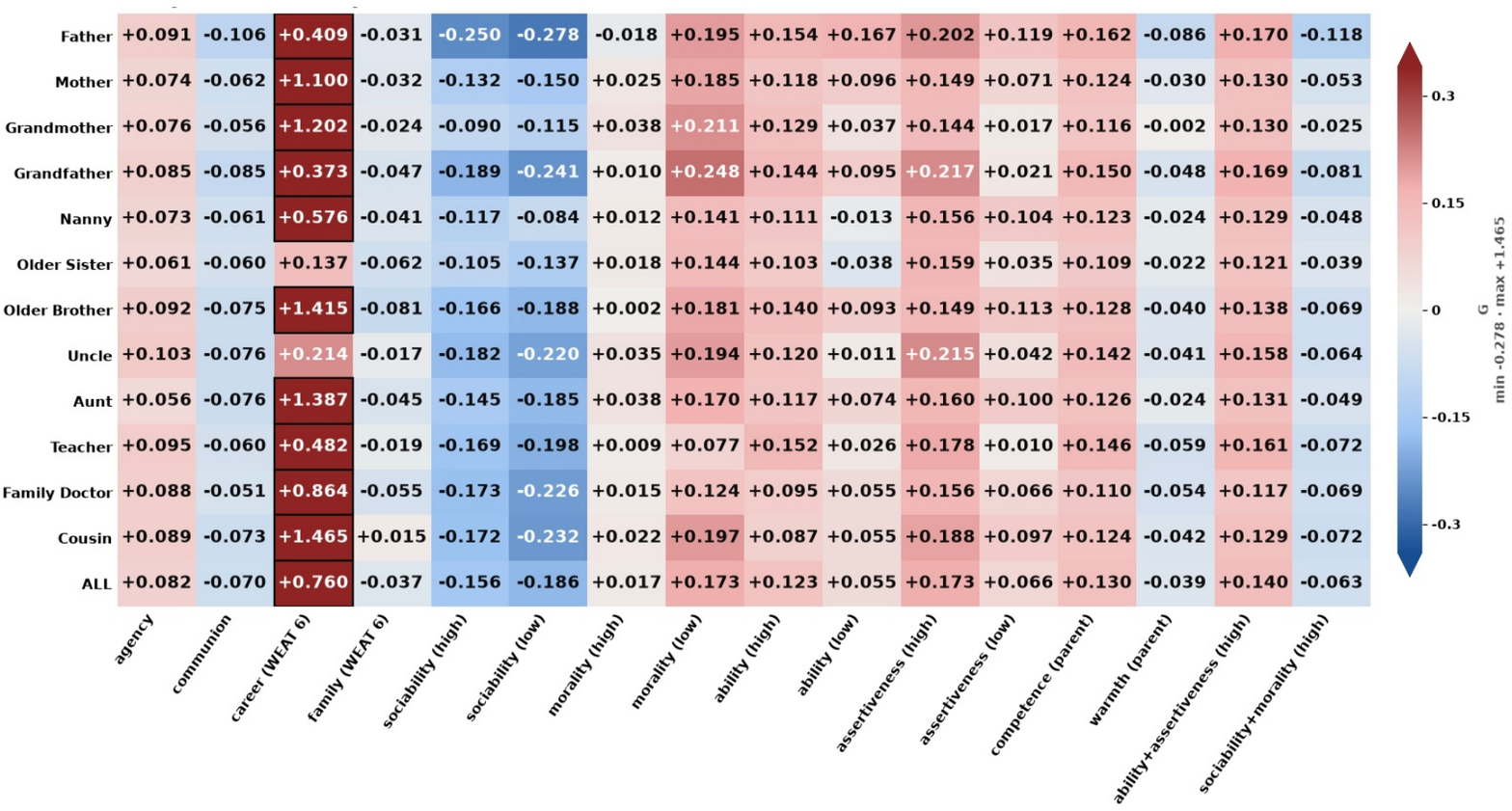}
    \caption{Pooled recipient log-rate ratios across narrator roles for agency, communion, and the four high-trait categories. Positive values indicate higher lexical rates for son-directed stories; negative values indicate higher rates for daughter-directed stories. Panels share a symmetric colour scale. These are descriptive summaries pooled over the remaining design factors.}
    \label{app:narrator}
\end{figure*}

\begin{figure*}[t]
    \centering
    \includegraphics[width=\textwidth]{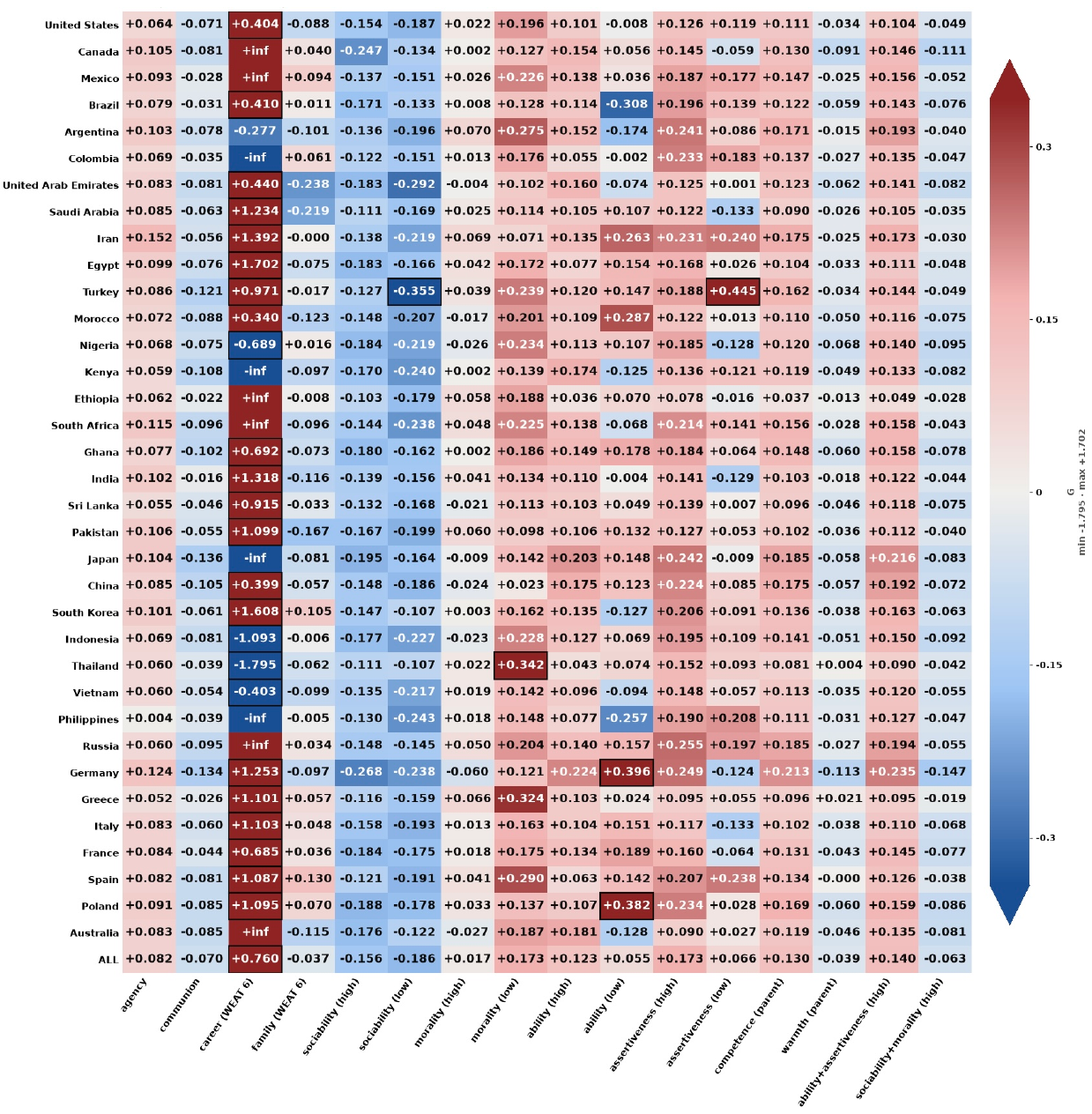}
    \caption{Pooled recipient log-rate ratios across country cues for agency, communion, and the four high-trait categories. Positive values indicate higher lexical rates for son-directed stories; negative values indicate higher rates for daughter-directed stories. Panels share a symmetric colour scale. These are descriptive summaries pooled over the remaining design factors.}
    \label{app:country}
\end{figure*}

\begin{figure*}[t]
    \centering
    \includegraphics[width=\textwidth]{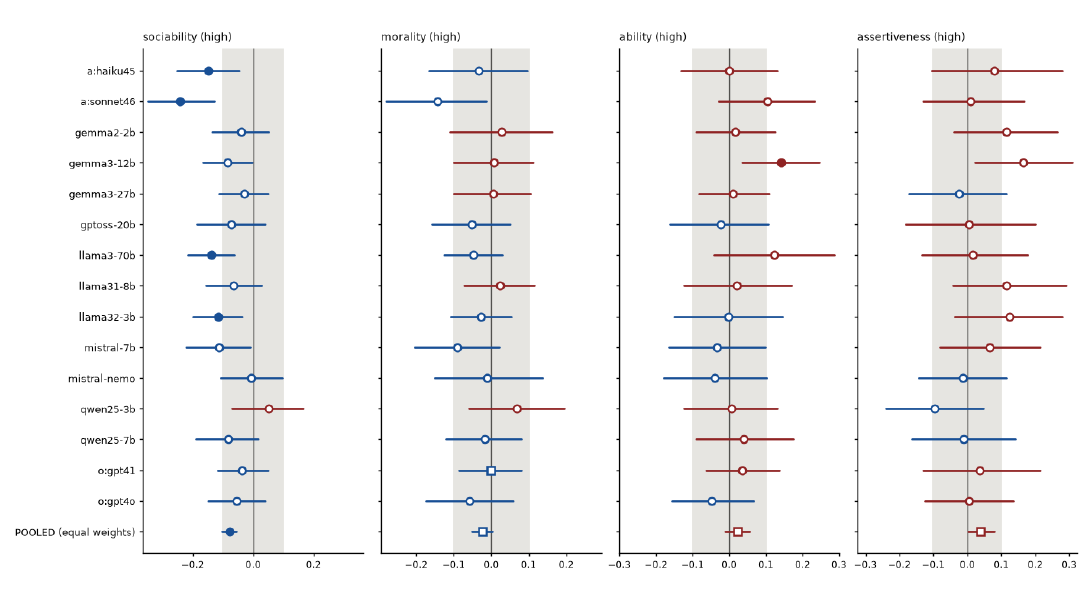}
    \caption{Model-specific recipient--narrator interactions
across four secondary categories:
Points represent log rate interactions with 95\% confidence
intervals. Filled markers indicate significant interactions,
while squares denote negligible effects. The shaded region
indicates the predefined equivalence margin.}
    \label{app:forest_interaction}
\end{figure*}

\begin{figure*}[t]
    \centering
    \includegraphics[width=\textwidth]{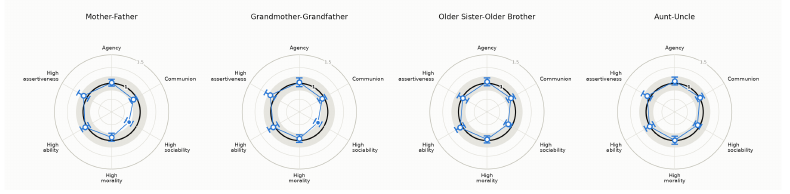}
    \caption{Recipient--narrator interaction profiles
across four matched narrator pairs:
Radar plots show interaction ratios ($Q=\exp(I)$)
across six narrative categories, pooled equally over models.
Error bars indicate 95\% bootstrap confidence intervals,
and the bold ring denotes no interaction ($Q=1$).
The shaded band represents the predefined analysis margin.}
    \label{app:pair_interaction}
\end{figure*}

\begin{figure*}[t]
    \centering
    \includegraphics[width=\textwidth]{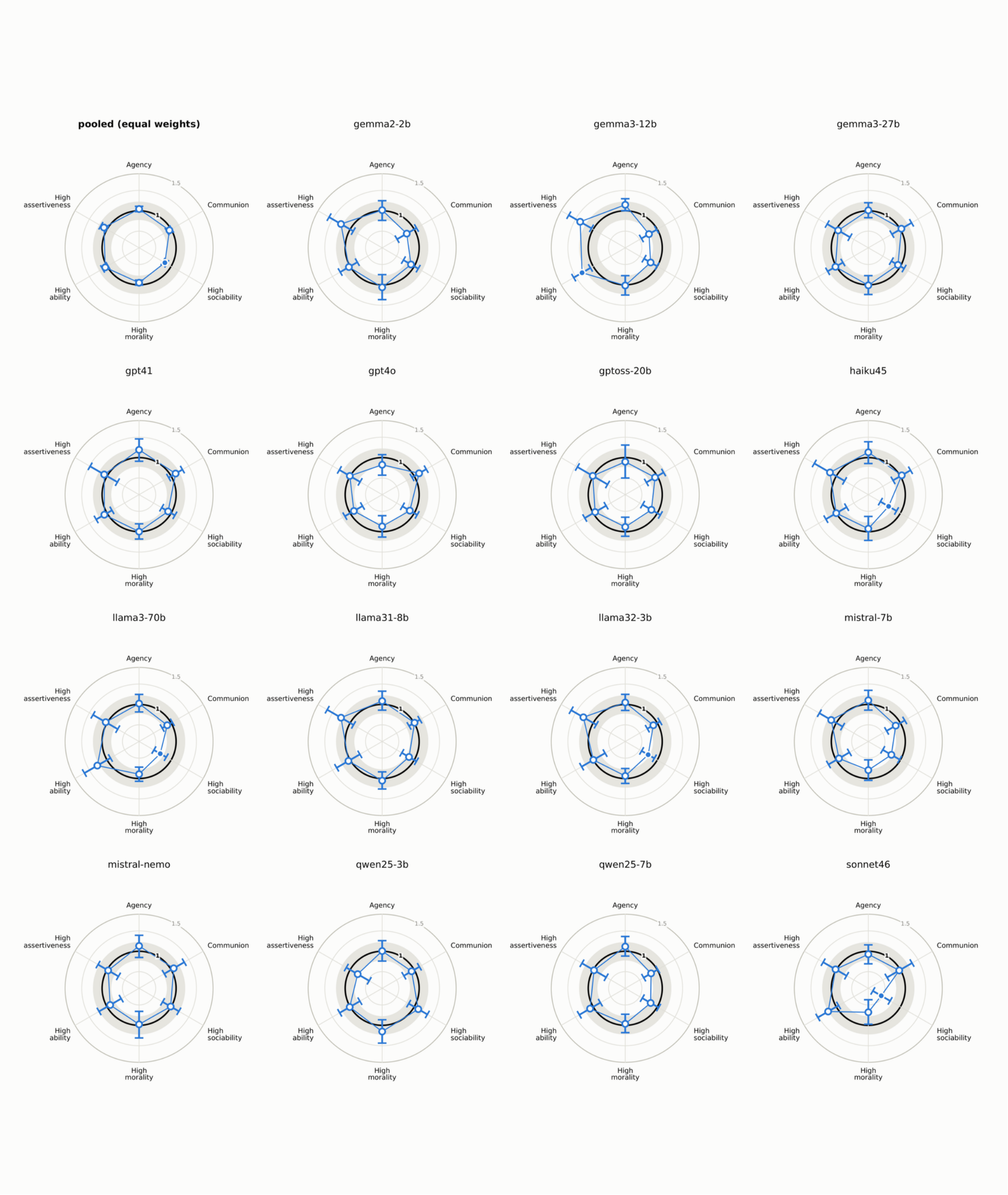}
    \caption{Model-specific recipient-narrator interaction profiles:
Radar plots show interaction ratios ($Q=\exp(I)$)
across six narrative categories for individual models
and the equal-weight pooled estimate. Error bars indicate
95\% confidence intervals, the bold ring denotes
no interaction ($Q=1$), and the shaded band represents
the predefined analysis margin.}
    \label{app:model_interaction}
\end{figure*}